\documentclass[runningheads]{llncs}

\usepackage{eccv}

\usepackage{eccvabbrv}

\usepackage{graphicx}
\usepackage{booktabs}

\usepackage[accsupp]{axessibility}  

\usepackage{hyperref}

\usepackage{orcidlink}

\begin{document}

\title{SAM3Dual: A 3rd Place Solution to the MOSEv2 Track, 8th LSVOS Challenge} 

\titlerunning{SAM3Dual: 3rd Place Solution to MOSEv2}

\author{
JeongRae Kim\inst{1} \and
Chaehyun Kim\inst{1} \and
Changwon Lim\inst{1}\thanks{Corresponding author.}
}

\authorrunning{J.~Kim et al.}

\institute{
Chung-Ang University, Seoul, Republic of Korea\\
\email{\{kjk632,rea828,clim\}@cau.ac.kr}
}

\maketitle

\begin{abstract}
We present \textbf{SAM3Dual}, our third-place solution to the MOSEv2
track of the 8th Large-scale Video Object Segmentation (LSVOS)
Challenge at ECCV 2026.
SAM3Dual is a training-free inference extension of pretrained SAM 3
that explicitly separates temporal memory into a short-term branch for
recent observations and a long-term branch for interval-sampled
historical representations.
The two memory responses are combined using a deterministic
sequence-relative fusion schedule and conservatively modulated by the
previous-frame object confidence.
All pretrained SAM 3 parameters remain frozen, requiring no
task-specific training, fine-tuning, test-time training, or online
parameter optimization.
The complete system achieved an official
$\mathcal{J}\&\mathcal{F}$ score of \textbf{64.37} and ranked
\textbf{third} in the MOSEv2 track.
This result highlights the potential of reorganizing temporal memory
entirely at inference time to obtain competitive long-term VOS
performance while preserving the pretrained model.
\end{abstract}

\section{Introduction}
\label{sec:introduction}

Video object segmentation (VOS) aims to delineate one or more target
objects throughout a video sequence given their annotations in the
initial frame.
Despite substantial progress, reliable propagation over long and
complex sequences remains challenging because targets may undergo rapid
motion, deformation, scale variation, occlusion, disappearance, and
reappearance.
These difficulties become particularly severe when visually similar
objects coexist in cluttered scenes, where prediction errors can
accumulate over time and eventually cause object drift or identity
switches.
Long-term VOS therefore requires a model to adapt to recent motion and
appearance changes while retaining sufficiently stable historical
information to preserve target identity over extended temporal gaps.

The MOSEv2 track of the 8th Large-scale Video Object Segmentation
(LSVOS) Challenge at ECCV 2026 evaluates semi-supervised VOS under such
challenging long-term conditions, including frequent occlusions,
object interactions, disappearance, reappearance, and substantial
appearance variation~\cite{MOSEv2,lsvos2026mosev2}.
This paper describes our submission to the MOSEv2 track.

Promptable segmentation foundation models provide a strong basis for
general-purpose video segmentation.
SAM 3 extends prompt-based segmentation to video and propagates target
information through a memory-based tracking mechanism
~\cite{carion2025sam3segmentconcepts}.
However, long-term propagation creates an inherent tension in temporal
memory organization.
Recent observations are highly relevant to the current target state
and facilitate adaptation to local changes, but they are derived from
propagated predictions and may contain accumulated errors.
Temporally distant observations can preserve earlier target
representations that remain useful after occlusion or disappearance,
but they may be less representative of the target's current
appearance.
A single streaming memory must therefore balance
\emph{recent adaptation} and \emph{historical identity preservation}.

Prior work has demonstrated the importance of structured temporal
memory in long-term VOS~\cite{cheng2022xmem}, while SAM2Long shows that
the memory organization of a pretrained segmentation model can be
modified entirely at inference time without updating its parameters
~\cite{ding2024sam2long}.
Motivated by these observations, we investigate whether recent and
temporally distant information can be explicitly separated within a
single frozen SAM 3 propagation pathway.

To this end, we propose \textbf{SAM3Dual}, a training-free inference
extension of SAM 3 based on explicit temporal memory separation.
SAM3Dual maintains a short-term branch containing recent observations
and a long-term branch containing interval-sampled historical
representations.
Both branches use the same frozen pretrained SAM 3 memory-attention
operation and differ only in the temporal composition of their memory
banks.
Their responses are combined using a deterministic sequence-relative
fusion schedule and conservatively modulated using the previous-frame
object confidence.
All pretrained SAM 3 parameters remain frozen, and the method requires
no task-specific training, fine-tuning, test-time training, or online
parameter optimization.

The complete SAM3Dual system achieved an official
$\mathcal{J}\&\mathcal{F}$ score of \textbf{64.37} and ranked
\textbf{third} in the MOSEv2 track of the 8th LSVOS Challenge.
Because the challenge submission was evaluated as a complete system,
we interpret this result as evidence of the competitiveness of the
overall training-free inference design rather than as a
component-wise performance claim.

Our main contributions are summarized as follows:
\begin{itemize}
    \item We introduce \textbf{SAM3Dual}, a training-free inference
    framework that separates recent and temporally dispersed
    historical observations into short-term and long-term memory
    branches while keeping all pretrained SAM 3 parameters frozen.

    \item We combine the two memory responses using a deterministic
    sequence-relative temporal fusion schedule together with bounded
    confidence-guided modulation, introducing no additional learnable
    parameters.

    \item We report the resulting system as our official MOSEv2
    challenge submission, which achieved an official
    $\mathcal{J}\&\mathcal{F}$ score of \textbf{64.37} and ranked
    \textbf{third}.
\end{itemize}

\section{Related Work}
\label{sec:related_work}

\subsection{Memory-Based Video Object Segmentation}

Memory-based propagation is a dominant paradigm in semi-supervised
video object segmentation (VOS).
STM~\cite{oh2019stm} introduced space-time memory matching, enabling
current-frame features to retrieve information from previously
segmented frames.
Subsequent methods improved temporal correspondence and multi-object
propagation: STCN~\cite{cheng2021stcn} enhanced space-time
correspondence and memory utilization, while AOT~\cite{yang2021aot}
and DeAOT~\cite{yang2022deaot} developed transformer-based
hierarchical propagation for multi-object VOS.

Long-term VOS places additional emphasis on how historical
observations are organized and retained.
XMem~\cite{cheng2022xmem} introduced sensory, working, and long-term
memory stores, demonstrating the benefit of assigning different
temporal roles to different memory representations.
More recent approaches have investigated complementary aspects of
memory design.
Cutie~\cite{cheng2024cutie} incorporates object-level memory reading
to reduce ambiguity from distractors, whereas
RMem~\cite{zhou2024rmem} shows that restricting the memory bank can
avoid redundant historical information and improve memory retrieval.

These studies collectively indicate that long-term VOS depends not
only on retaining historical information but also on how that
information is organized, selected, and retrieved.
SAM3Dual follows this general motivation but differs from conventional
memory-based VOS architectures in that it does not train a new memory
representation or temporal module.
Instead, it preserves the pretrained SAM 3 memory-attention mechanism
and reorganizes its inputs according to temporal role.
The short-term and long-term branches therefore differ in their
temporal memory composition rather than in separately learned
encoders or attention parameters.

\subsection{Foundation Models and Training-Free Long-Term VOS}

The Segment Anything Model (SAM)~\cite{kirillov2023sam} established
large-scale promptable image segmentation, and SAM 2
~\cite{ravi2024sam2} extended this paradigm to images and videos using
a streaming-memory architecture.
SAM 3~\cite{carion2025sam3segmentconcepts} further extends the
Segment Anything family to concept-conditioned detection,
segmentation, and tracking while retaining memory-based temporal
propagation for video.
These foundation models provide strong pretrained representations, but
long-video segmentation remains sensitive to the organization of
temporal memory, particularly under occlusion, disappearance,
reappearance, and accumulated prediction errors.

Training-free approaches seek to exploit these pretrained models
without additional parameter updates.
Most closely related to our work, SAM2Long
~\cite{ding2024sam2long} addresses long-video error accumulation in
SAM 2 by maintaining multiple candidate propagation histories in a
memory tree and selecting promising paths according to cumulative
confidence.
This demonstrates that modifying memory usage at inference time can
improve the robustness of a frozen video segmentation model.

SAM3Dual shares the training-free objective of SAM2Long but adopts a
different inference strategy.
Rather than maintaining multiple competing propagation paths,
SAM3Dual retains a single SAM 3 propagation pathway and explicitly
separates its temporal memory into short-term and long-term branches.
The two responses are combined through a deterministic
sequence-relative fusion schedule and bounded confidence-guided
modulation.
Thus, SAM3Dual focuses on reorganizing the temporal composition of
memory within a single frozen SAM 3 inference process rather than
searching over multiple segmentation hypotheses.

\section{Method}
\label{sec:method}

\subsection{Overview}
\label{sec:overview}

SAM3Dual extends pretrained SAM 3 with a training-free dual-memory
mechanism for long-term video object segmentation.
Given the target annotation in the first frame, SAM 3 initializes the
target representation and propagates the target mask through the
remaining frames.
All pretrained SAM 3 parameters remain frozen throughout inference.

\begin{figure*}[t]
\centering
\includegraphics[width=\textwidth]{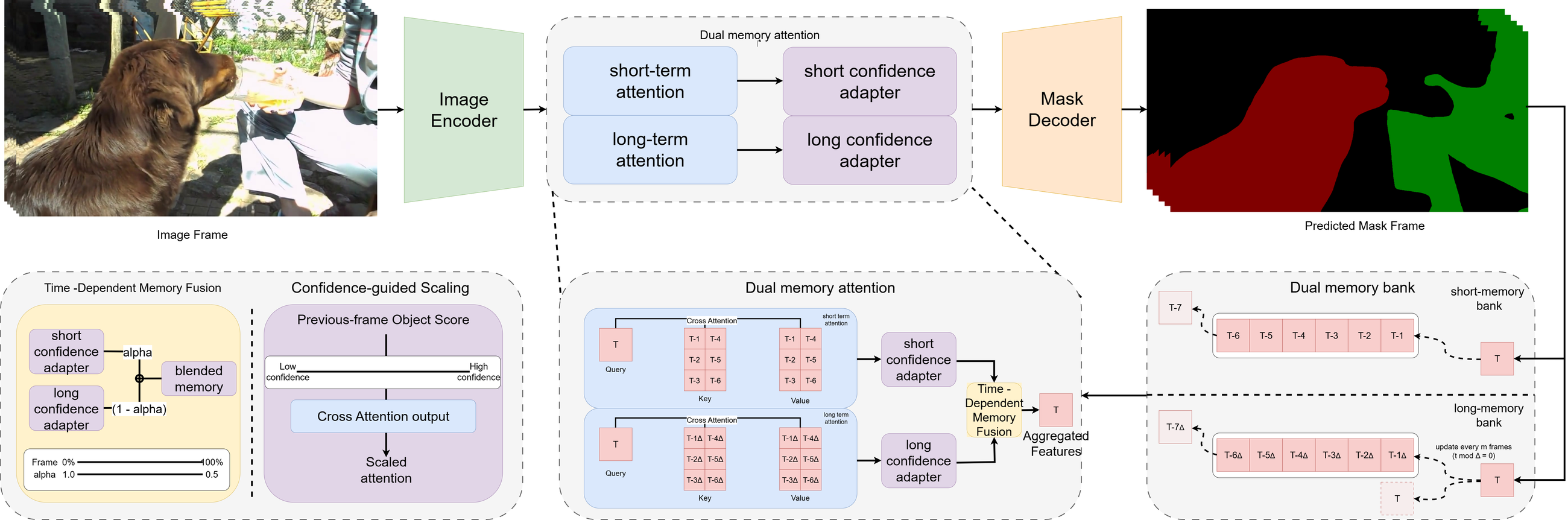}
\caption{
Overview of SAM3Dual.
Current-frame features independently attend to short-term memory
containing recent observations and long-term memory containing
interval-sampled historical observations.
Their responses are confidence-modulated and combined using a
sequence-relative temporal fusion schedule before being passed to the
frozen SAM 3 prediction pathway.
}
\label{fig:sam3dual_architecture}
\end{figure*}

SAM3Dual explicitly separates temporal information into short-term and
long-term memory branches.
The short-term branch retains recent observations, whereas the
long-term branch maintains temporally dispersed historical
representations.
Current-frame features independently retrieve information from both
branches using the same frozen SAM 3 memory-attention operation.
The resulting responses are modulated using the previous-frame object
confidence and combined through a deterministic sequence-relative
fusion schedule.
The overall architecture is illustrated in
Fig.~\ref{fig:sam3dual_architecture}.

\subsection{Dual-Memory Architecture}
\label{sec:dual_memory}

Let $I_t$ denote the input frame at time $t$, and let $Q_t$ denote the
current-frame representation used as the query for memory attention.
SAM3Dual maintains a short-term memory bank $\mathcal{M}^{S}_t$ and a
long-term memory bank $\mathcal{M}^{L}_t$.
In both branches, the representation derived from the first-frame
ground-truth annotation is permanently retained as a reference to the
initially specified target.
Each branch additionally maintains up to six temporal memory entries,
resulting in a bounded $1+6$ memory structure.
The branches differ in the temporal composition of these additional
entries.

\paragraph{Short-term memory.}
The short-term branch follows the streaming-memory mechanism used in
our SAM 3 inference configuration and retains recent temporal
observations.
Let $K^{S}_t$ and $V^{S}_t$ denote the keys and values retrieved from
$\mathcal{M}^{S}_t$.
The short-term memory response is

\begin{equation}
O^{S}_t =
\operatorname{CrossAttn}_{\theta}
\left(Q_t,K^{S}_t,V^{S}_t\right),
\label{eq:short_memory}
\end{equation}

where $\operatorname{CrossAttn}_{\theta}$ denotes the pretrained
SAM 3 memory-attention operation with frozen parameters $\theta$.

\paragraph{Long-term memory.}
The long-term branch uses the same bounded $1+6$ structure but
maintains temporally dispersed historical observations.
The first-frame reference is retained throughout the sequence, while
the additional entries are selected at fixed temporal intervals.
Let $K^{L}_t$ and $V^{L}_t$ denote the corresponding keys and values.
The long-term response is

\begin{equation}
O^{L}_t =
\operatorname{CrossAttn}_{\theta}
\left(Q_t,K^{L}_t,V^{L}_t\right).
\label{eq:long_memory}
\end{equation}

Equations~\eqref{eq:short_memory} and~\eqref{eq:long_memory} use the
same frozen attention parameters.
Thus, the distinction between the branches arises from their temporal
memory composition rather than from separately learned encoders or
attention mechanisms.

\subsection{Confidence-Guided Memory Modulation}
\label{sec:confidence_modulation}

SAM3Dual uses the object confidence predicted at the previous frame to
conservatively modulate the magnitude of the memory responses.
Let $z_{t-1}$ denote the previous-frame object-confidence logit.
We define

\begin{equation}
c_{t-1}=\sigma(z_{t-1}), \qquad
s_t=s_{\min}+(s_{\max}-s_{\min})c_{t-1},
\label{eq:confidence_scale}
\end{equation}

where $c_{t-1}\in[0,1]$ and the submitted system uses
$s_{\min}=0.9$ and $s_{\max}=1.1$.
The two memory responses are then scaled as

\begin{equation}
\widetilde{O}^{S}_t=s_tO^{S}_t,
\qquad
\widetilde{O}^{L}_t=s_tO^{L}_t.
\label{eq:scaled_memory}
\end{equation}

High confidence therefore slightly amplifies the memory-derived
responses, whereas low confidence attenuates them.
The narrow range $[0.9,1.1]$ keeps the modulation close to the
pretrained response.
Because the same $s_t$ is applied to both branches, confidence
modulation does not change their relative weighting; it only controls
the overall magnitude of the memory contribution.

\subsection{Sequence-Relative Temporal Fusion}
\label{sec:temporal_fusion}

SAM3Dual combines the two memory responses using a deterministic
temporal schedule.
Let $N$ denote the total number of frames and let $\alpha$ denote the
minimum short-term weight.
At frame $t$, the short-term fusion coefficient is

\begin{equation}
g_t =
1-(1-\alpha)\frac{t}{N-1},
\qquad t\in\{0,\ldots,N-1\}.
\label{eq:fusion_gate}
\end{equation}

We use $\alpha=0.5$, so $g_t$ decreases linearly from $1.0$ to $0.5$
over the sequence, while the long-term weight $1-g_t$ increases from
$0$ to $0.5$.
The fused memory response is

\begin{equation}
O_t =
g_t\widetilde{O}^{S}_t
+
(1-g_t)\widetilde{O}^{L}_t.
\label{eq:memory_fusion}
\end{equation}

Thus, $g_t$ determines the relative contribution of recent and
historical information, whereas the confidence factor $s_t$ globally
modulates their memory response.
Normalizing time by $N-1$ allows the same schedule to be applied to
videos of different lengths without a learned gating module.

The formulation assumes that the total sequence length $N$ is known
before propagation, which is satisfied by the offline MOSEv2 challenge
setting.
Strictly online video would require a causal fusion schedule that does
not depend on $N$.

\subsection{Memory Update and Training-Free Inference}
\label{sec:memory_update}

After predicting the target mask at frame $t$, SAM 3 generates a
memory representation associated with the current prediction.
The first-frame ground-truth representation remains fixed in both
memory banks throughout the sequence.

The short-term branch follows the SAM 3 streaming update policy and
retains up to six recent temporal entries.
The long-term branch uses the same bounded $1+6$ memory structure but
considers historical observations at intervals of $\Delta$ frames.
For the submitted system, $\Delta=10$.
Consequently, the long-term bank contains the persistent first-frame
reference together with up to six temporally dispersed historical
entries, preventing memory size from increasing with video length.

No model parameters are updated during this procedure.
The pretrained image encoder, memory encoder, memory-attention modules,
and mask decoder remain frozen.
SAM3Dual therefore requires no task-specific training, fine-tuning,
test-time training, or online parameter optimization.

\section{Experiments}
\label{sec:experiments}

\subsection{Evaluation Protocol}
\label{sec:evaluation_protocol}

We evaluate SAM3Dual on the MOSEv2 track of the 8th Large-scale Video
Object Segmentation (LSVOS) Challenge at ECCV 2026
~\cite{MOSEv2,lsvos2026mosev2}.
MOSEv2 evaluates semi-supervised VOS under challenging long-term
conditions including occlusion, disappearance, reappearance, object
interactions, and substantial appearance variation.

Following the official protocol, each target is initialized from the
ground-truth annotation in the first frame, after which its mask is
propagated through the remaining frames without additional
ground-truth supervision.
Performance is measured using the official
$\mathcal{J}\&\mathcal{F}$ metric.
The metric averages region similarity $\mathcal{J}$ and contour
accuracy $\mathcal{F}$.
All leaderboard scores reported in this paper are official challenge
results.
Our final system was submitted under the account name
\texttt{kjeong}.

\subsection{Implementation Details}
\label{sec:implementation}

We use the publicly released pretrained SAM 3 checkpoint
\texttt{facebook/sam3}.
Inference is performed on a single NVIDIA GeForce RTX 4090 GPU using
BF16 mixed precision.
All pretrained SAM 3 parameters remain frozen, and SAM3Dual requires
no task-specific training, fine-tuning, test-time training, or online
parameter optimization.

The inference configuration used for the official submission is
summarized in Table~\ref{tab:inference_settings}.
The memory organization and fusion mechanisms are described in
Sec.~\ref{sec:method}.
All settings were fixed for the submitted system and are not claimed
to be individually optimal.

\begin{table}[h]
\centering
\small
\caption{Inference configuration of the official SAM3Dual submission.}
\label{tab:inference_settings}
\begin{tabular}{ll}
\hline
\textbf{Setting} & \textbf{Value} \\
\hline
Base model & \texttt{facebook/sam3} \\
Hardware & 1$\times$ NVIDIA GeForce RTX 4090 \\
Precision & BF16 mixed precision \\
Training / adaptation & None \\
Short-term memory & $1+6$ entries \\
Long-term memory & $1+6$ entries \\
First-frame GT reference & Always retained \\
Long-term interval $\Delta$ & 10 frames \\
Fusion parameter $\alpha$ & 0.5 \\
Confidence range & $[0.9,1.1]$ \\
\hline
\end{tabular}
\end{table}

\section{Results and Analysis}
\label{sec:results}

\subsection{Official Challenge Results}
\label{sec:official_results}

Table~\ref{tab:leaderboard} presents the top-three entries on the
official leaderboard of the MOSEv2 track.
SAM3Dual, submitted under the account name \texttt{kjeong}, achieved
an official $\mathcal{J}\&\mathcal{F}$ score of \textbf{64.37} and
ranked \textbf{third}.
The first- and second-ranked submissions achieved scores of 69.82 and
66.20, respectively.

\begin{table}[t]
\centering
\small
\caption{Top-three entries on the official MOSEv2 leaderboard of the
8th LSVOS Challenge at ECCV 2026.}
\label{tab:leaderboard}
\begin{tabular}{clc}
\hline
\textbf{Rank} & \textbf{Team / Account} &
$\boldsymbol{\mathcal{J}\&\mathcal{F}}\uparrow$ \\
\hline
1 & HITsz-Dragon & 69.82 \\
2 & mmm & 66.20 \\
3 & AISTAT (\texttt{kjeong}) & \textbf{64.37} \\
\hline
\end{tabular}
\end{table}

The reported score corresponds to the complete training-free
SAM3Dual system evaluated under the official challenge protocol.
Because the original SAM 3 configuration and individual SAM3Dual
components were not separately evaluated under an identical official
setting, we do not attribute the leaderboard result to any individual
component.
The following analysis therefore focuses on representative qualitative
behavior of the complete submitted system.

\subsection{Qualitative Analysis}
\label{sec:qualitative_analysis}

\begin{figure*}[t]
\centering
\includegraphics[width=0.85\textwidth]{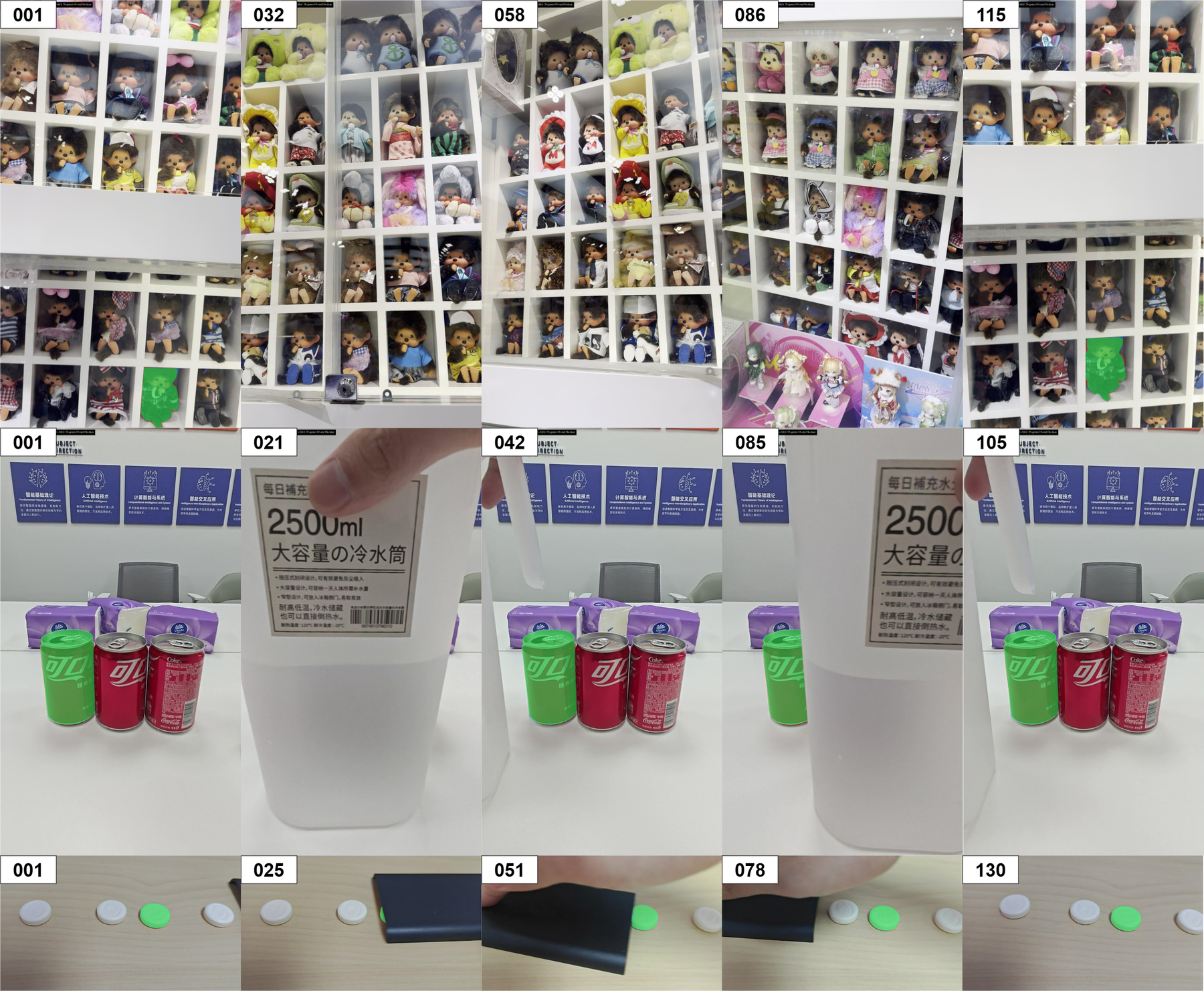}
\caption{Representative successful cases of SAM3Dual on the MOSEv2
training set. Green, red, and blue indicate true positives, false
positives, and false negatives, respectively.}
\label{fig:positive}
\end{figure*}

Figure~\ref{fig:positive} shows representative successful cases on the
MOSEv2 training set.
Green, red, and blue indicate true-positive, false-positive, and
false-negative regions, respectively.
The examples illustrate stable target propagation under challenging
temporal changes including motion, partial occlusion, disappearance,
and appearance variation.

These examples are consistent with the intended complementary roles of
the two memory branches.
Recent short-term observations provide information closely related to
the current target state, while temporally dispersed historical
memories remain available when recent observations become less
informative after occlusion or disappearance.
By exposing both temporal ranges to the frozen prediction pathway,
SAM3Dual can maintain target masks across temporally separated target
appearances.

The examples should not be interpreted as component-wise evidence,
since the two memory branches and fusion mechanism were not evaluated
independently.
Rather, they illustrate representative conditions under which the
complete SAM3Dual system produced visually consistent segmentation.

\subsection{Failure Case Analysis}
\label{sec:failure_analysis}

\begin{figure*}[t]
\centering
\includegraphics[width=0.85\textwidth]{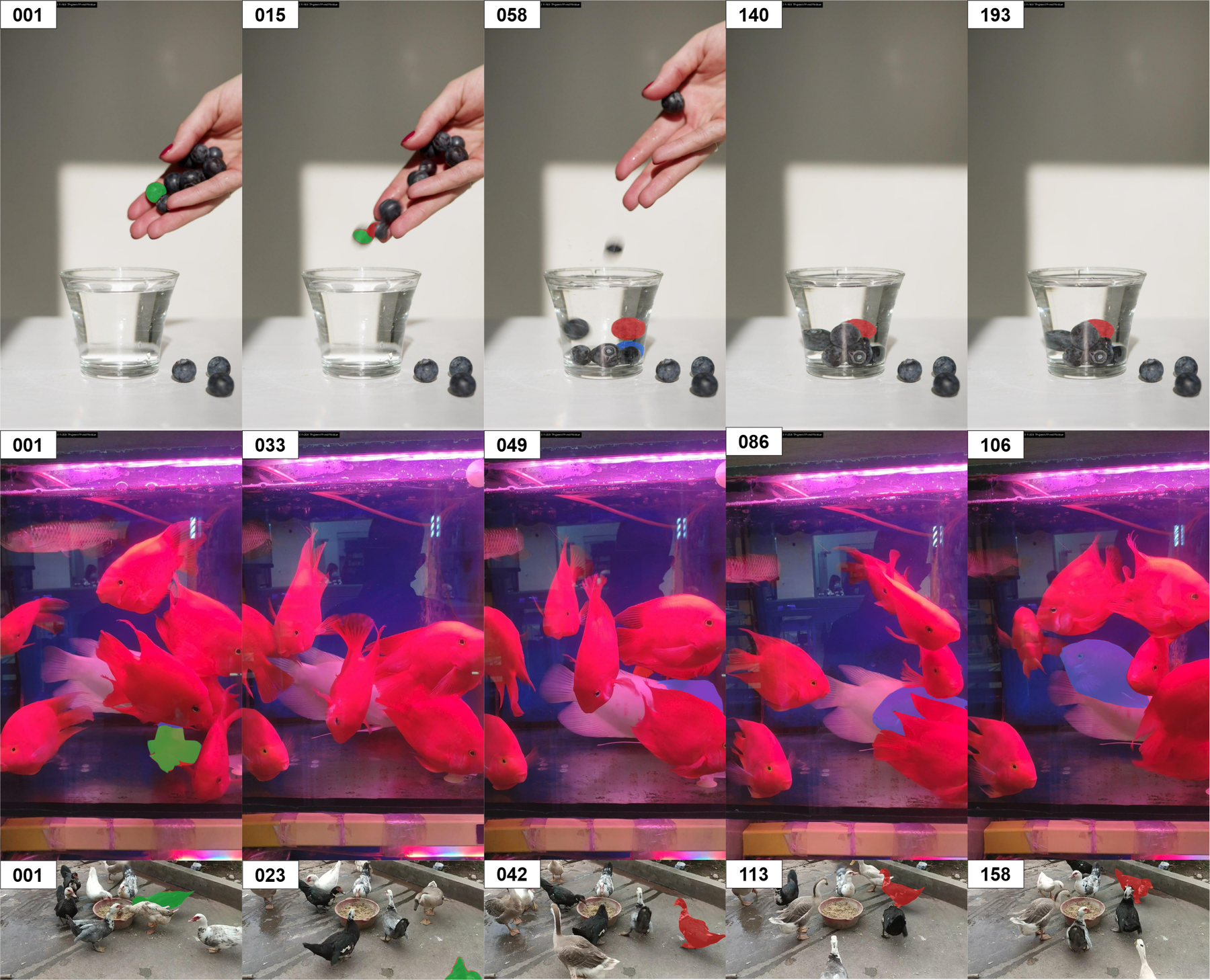}
\caption{Representative failure cases on the MOSEv2 training set,
primarily involving visually similar distractors, large target
displacement, and reappearance. Colors follow
Fig.~\ref{fig:positive}.}
\label{fig:negative}
\end{figure*}

Despite its competitive challenge result, SAM3Dual still exhibits
identity errors in difficult long-term sequences.
Figure~\ref{fig:negative} illustrates two recurring failure patterns.

\paragraph{Appearance ambiguity.}
When the target and nearby distractors have similar appearances,
historical appearance representations may not uniquely identify the
correct object.
Memory retrieval can consequently associate the current observation
with a visually plausible distractor, producing false positives on the
incorrect object together with false negatives on the actual target.

\paragraph{Large displacement and reappearance.}
A second failure pattern occurs when the target disappears or becomes
occluded and later reappears at a substantially different position.
SAM3Dual reorganizes temporal appearance memory but does not introduce
an explicit mechanism for long-range spatial correspondence or target
relocation.
When large spatial displacement coincides with a visually similar
distractor, the stored target representation can therefore be
associated with the wrong object.

These cases highlight an important limitation of appearance-based
temporal memory:
\emph{preserving historical appearance information is not equivalent
to preserving object identity}.
This distinction motivates stronger identity-aware and
uncertainty-aware memory mechanisms, which are discussed further in
Sec.~\ref{sec:discussion}.

\section{Discussion and Limitations}
\label{sec:discussion}

\subsection{Interpretation of the Challenge Result}

SAM3Dual achieved an official
$\mathcal{J}\&\mathcal{F}$ score of \textbf{64.37} and ranked
\textbf{third} in the MOSEv2 track.
The result demonstrates the feasibility of a competitive
training-free long-term VOS system that reorganizes temporal memory
at inference time while keeping the pretrained SAM 3 parameters
frozen.

However, the result should be interpreted at the complete-system
level.
The original SAM 3 configuration and individual SAM3Dual components
were not evaluated under an identical official setting, and no
controlled ablation study was conducted for the submitted system.
We therefore do not attribute the leaderboard score to any individual
component or claim that the submitted hyperparameters are optimal.
Rather, the result establishes the competitiveness of the complete
training-free inference configuration used for the challenge.

\subsection{Computational and Practical Considerations}

Both memory branches use a bounded $1+6$ structure, preventing memory
usage from growing indefinitely with sequence length.
SAM3Dual nevertheless requires an additional long-term memory bank and
a second memory-attention operation relative to the original
single-branch SAM 3 propagation pathway.
Confidence modulation and temporal fusion themselves require only
scalar rescaling and weighted combination, but the additional memory
retrieval necessarily introduces extra inference-time computation and
memory usage.

We did not perform controlled measurements of runtime, throughput, or
peak GPU memory against the original SAM 3 configuration and therefore
make no quantitative efficiency claim.
Such measurements would be useful for characterizing the
performance--efficiency trade-off of the dual-memory design.

\subsection{Limitations and Future Directions}

\paragraph{Fixed memory configuration.}
The submitted system uses a fixed memory capacity ($1+6$), a fixed
long-term sampling interval ($\Delta=10$), and a predefined
sequence-relative fusion schedule with $\alpha=0.5$.
These deterministic choices provide bounded and predictable inference,
but they cannot adapt to sequence-specific variations in motion,
visibility, appearance change, or scene complexity.
Adaptive memory selection, content-dependent memory replacement, and
event-aware fusion could provide more flexible temporal processing
while retaining the training-free design.

\paragraph{Limited confidence use and memory errors.}
The current confidence mechanism applies the same bounded factor
$s_t\in[0.9,1.1]$ to both memory branches and therefore modulates only
the overall magnitude of the memory response rather than their
relative reliability.
Moreover, although the first-frame ground-truth representation is
permanently retained, the remaining temporal memories are generated
from propagated predictions.
Incorrect predictions can therefore introduce unreliable
representations into subsequent memory.
Future designs could use confidence or consistency signals not only
for response scaling but also for branch weighting, memory insertion,
and memory replacement.

\paragraph{Identity ambiguity.}
The failure cases in Sec.~\ref{sec:failure_analysis} show that
preserving historical appearance information does not necessarily
preserve target identity.
When a target reappears after occlusion or disappearance at a
substantially different location in the presence of visually similar
objects, stored appearance representations may match an incorrect
distractor.
SAM3Dual does not explicitly model long-range spatial correspondence,
target relocation, or re-identification.
Training-free identity matching or spatial-consistency mechanisms are
therefore promising directions for improving robustness under severe
appearance ambiguity and displacement.

\paragraph{Offline temporal fusion.}
The current fusion schedule depends on the normalized temporal
position $t/(N-1)$ and therefore assumes that the total sequence length
$N$ is known before propagation.
This is appropriate for the offline MOSEv2 challenge setting, but a
strictly online application would require a causal or event-driven
fusion rule independent of the final sequence length.

Overall, SAM3Dual should be viewed as a challenge solution
demonstrating the feasibility of bounded dual-memory organization in a
frozen SAM 3 inference pipeline rather than as a comprehensive study
of long-term memory design.
Controlled comparisons of memory capacities, sampling strategies,
fusion mechanisms, and confidence usage remain important directions
for future evaluation.

\section{Conclusion}
\label{sec:conclusion}

We presented \textbf{SAM3Dual}, our third-place solution to the MOSEv2
track of the 8th LSVOS Challenge at ECCV 2026.
SAM3Dual reorganizes the temporal memory of pretrained SAM 3 into
short-term and long-term branches and combines their responses through
deterministic temporal fusion and confidence-guided modulation.
The entire framework operates at inference time with all pretrained
SAM 3 parameters frozen and requires no task-specific training,
fine-tuning, test-time training, or online parameter optimization.
The complete system achieved an official
$\mathcal{J}\&\mathcal{F}$ score of \textbf{64.37} and ranked
\textbf{third} in the MOSEv2 track.

These results demonstrate the feasibility of training-free temporal
memory reorganization for competitive long-term VOS.
At the same time, qualitative failures under appearance ambiguity,
large displacement, and reappearance indicate that preserving
historical appearance alone is insufficient for reliable target
identity preservation.
Future work will investigate adaptive memory selection and fusion,
uncertainty-aware memory updates, and stronger identity-preserving
mechanisms while retaining the training-free design.


\bibliographystyle{splncs04}
\bibliography{main}

\end{document}